\documentclass[conference]{IEEEtran}
\IEEEoverridecommandlockouts
\usepackage[margin=2.5cm]{geometry}
\usepackage{amsmath, amssymb, amsthm}
\usepackage{booktabs}
\usepackage{hyperref}
\usepackage{xcolor}
\usepackage{algorithm}
\usepackage{algpseudocode}
\usepackage{graphicx}
\usepackage{cite}
\usepackage{textcomp}
\usepackage{pgfplots}
\pgfplotsset{compat=1.18}
\usepackage{subcaption}
\def\BibTeX{{\rm B\kern-.05em{\sc i\kern-.025em b}\kern-.08em
    T\kern-.1667em\lower.7ex\hbox{E}\kern-.125emX}}

\definecolor{highlight}{RGB}{30,100,200}
\definecolor{outerHBB}{HTML}{378ADD}
\definecolor{areaEquiv}{HTML}{888780}
\definecolor{gbbMarg}{HTML}{D85A30}
\definecolor{novelShape}{HTML}{1D9E75}

\begin{document}

\title{\textbf{Shape-Aware Oriented Bounding Box (OBB) to Horizontal Bounding Box (HBB) Conversion}\\
\thanks{Code and data are available on GitHub at \url{https://github.com/SkyServe-AI/obbtohbb} and archived on Zenodo at \url{https://doi.org/10.5281/zenodo.20775562}}}

\author{\IEEEauthorblockN{1\textsuperscript{st} Badha Rathna Sabhapathy}
\IEEEauthorblockA{ \textit{Consultant - Hyspace Technologies} \\
\textit{badharathna@gmail.com}\\
ORCID: 0009-0008-4672-1993
}
\and
\IEEEauthorblockN{2\textsuperscript{nd} Gotam Dahiya}
\IEEEauthorblockA{\textit{Hyspace Technologies} \\
\textit{gotam.dahiya@skyserve.ai}\\
ORCID: 0009-0009-7440-5813
}
\and
\IEEEauthorblockN{3\textsuperscript{rd} Vishesh Vatsal}
\IEEEauthorblockA{\textit{Hyspace Technologies} \\
\textit{vishesh@dfy-graviti.com}\\
ORCID: 0009-0008-2701-0010
}
}

\maketitle


\begin{abstract}
Accurate object detection in aerial and satellite imagery is dependent upon the bounding box representation. This is especially true for spatially oriented objects such as ships or aircrafts. Oriented Bounding Boxes (OBB) have a tighter fit and more robust non-max suppression compared to Horizontal Bounding Boxes (HBB), any current post-processing conversion from OBB to HBB either introduces excess empty and background space or removes data from the detection. This paper introduces a novel approach for a shape-aware OBB-to-HBB conversion for ship detection in remote sensing imagery. It leverages hull shape, hull fullness, and the bounding box orientation to produce a tighter axis-aligned HBB representation. The proposed method is benchmarked against three baselines methods for OBB-to-HBB conversion, Outer HBB which uses minimum and maximum, Area Equivalent HBB and GBB Marginalized HBB.\\

\end{abstract}


\section{Introduction}\label{INTRO}



With the spread of real world applications for satellite imagery, models required for classification, detection and segmentation have seen an increase in demand. With this increase, the data collected along with the annotations play a major role in creating a more accurate model. The more accurate the model, the better the results. In classification and segmentation, the output is straightforward, either classify an area or classify the individual pixels within the area to get a better understanding of the image. These models depend upon a threshold or the maximum values obtained from the model. Detections require the model to predict where the object is within the image. It tries to find the best bounding box for the detected object and present that as the result.

There are two main output formats for detection models, Oriented Bounding Boxes (OBB) and Horizontal Bounding Boxes (HBB). The differences between the two output formats stems from how the data is represented. Horizontal Bounding Boxes denote the objects bounds with the bounding box aligned with the image's axes. The two popular methods for this are, storing the center of the image along with height and width\cite{FasterRCNN} or represented with minimum and maximum corners of the bounding box\cite{CornerNet}. As tight as the bounding box can get, empty and background noise surrounding the object is introduced. This can cause the model to become confused, making assumptions based on the objects surroundings.

Object Oriented bounding boxes allow for minimal empty and background noise, as the bounding box tightly fits to the object's dimensions and orientation with respect to the image. Similar to HBB, there are two methods for representing an OBB, set of four corner coordinates\cite{DOTA-OBB} or using center of the object coordinates, length, width, and heading\cite{OBB-Angle}. An advantage of OBB over HBB is a more accurate non-max suppression (NMS)\cite{NMS} of detections returned by the model\cite{DualDet}. The orientation of the returned bounding box assists in the robustness of non-max suppression.

For post-processing after model prediction and NMS, conversion of OBB to HBB is generally done by using the minimum and maximum corners method. For a more tighter fit of HBB, a second approach of axis aligned bounding box based on orientation angle is used\cite{DOTA-OBB}. This paper introduces a novel approach for conversion of OBB to HBB for ship detections based on hull shape, hull fullness and orientation of bounding box with respect to the image. This produces a tighter fit for the HBB compared to other commonly used conversion methods.

\section{Related Work}\label{REL-WORK}

In this field, there is more research into conversion from the HBB to OBB using contour detection\cite{OBB-Contour} or segmentation masks\cite{OBBSegmentation}. To convert from OBB to HBB as explained in \ref{INTRO}, there are two standards methods for conversion. This section will explain in detail the current standards in detail.

As introduced in \cite{FasterRCNN} an HBB can be represented using the minimum and maximum coordinates of the box. This is a list of two coordinates represented as [(x1, y1), (x2, y2)]. This is the most basic HBB format, storing only length and width of the box. One issue with this type of bounding box is that the orientation or the actual length and width of the object is lost. This loss of information can result in requiring more data to train with for better detections. While converting from OBB to this format, the bounding box can be largely inflated, exaggerating the previously listed disadvantages. This is highly visible when creating shards for the detected object from the input image. For edge processing, this is disadvantageous for down-link of detected objects as the transmission size can increase with ever increasing detected OBBs and objects.

A second method is keeping the orientation angle and finding the intersection of the axes with the oriented bounding box from the center of the detected bounding box\cite{AngleOBBHBB}. This method compared to above aligns the HBB created more with the object than completely spaced out. Another advantage is that the orientation of the object is not completely lost. This is useful for detecting planes, ships, other vehicles and then classifying based on the shard extracted using the bounding box. One disadvantage while using angle, and length and width, some areas of the detected object can be cut out from the new bounding box made. This is can be detrimental to a classification model as well as losing data even before transmission. 
This method is also similar to using the midpoints of the OBB, but it cuts out too much data to be even useful for anything.

To maintain the detected object while providing the tightest fit for the bounding box, this paper proposes using the angle of the bounding box, the length and width of the bounding box with set parameters for hull shape to produce a horizontal bounding box which matches the ship's hull.

\section{Problem Statement}
\label{sec:problem}

In many ship detection pipelines, the model produces oriented bounding boxes because they represent rotation and aspect ratio more accurately. However, post-processing often requires axis-aligned horizontal bounding boxes because of evaluation protocols, annotation formats, or competition requirements.

This need arose in our maritime ship detection setting, where the baseline model produced oriented bounding boxes but the evaluation used horizontal bounding boxes. A direct conversion by enclosing the oriented box corners adds unnecessary background around the ship and lowers localization accuracy, which can reduce IoU-based performance and affect the final ranking.

To avoid this accuracy loss, we need a conversion method that:
\begin{itemize}
  \item takes an OBB (from a minimum rotated rectangle fit to the ship) as input,
  \item computes the tightest HBB that encloses the \emph{actual ship hull}, not just the OBB corners,
  \item preserves as much of the original geometric fidelity as possible, and
  \item does not require retraining the detection model.
\end{itemize}

Formally, given an OBB with half-length \(a = L/2\), half-width \(w = W/2\), and rotation angle \(\theta\) (the angle between the ship's long axis and the horizontal), we must compute an HBB centered at the OBB centroid \((\hat{x}, \hat{y})\) and aligned to the world axes, such that the HBB tightly encloses an inferred ship hull rather than the OBB extent.

\subsection{Hull Shape Model}
The ship cross-section is approximated by a \textbf{super-ellipse} (Lam\'{e} curve) in the ship's local frame:
\begin{equation}
  \frac{|x|^{q}}{a^{q}} + \frac{|y|^{q}}{b^{q}} = 1,
  \label{eq:superellipse}
\end{equation}
where
\begin{align*}
  a &= \frac{L}{2} \quad \text{(half-length from MRR)},\\
  b &= \frac{W}{2} \cdot f \quad \text{(half-width scaled by fullness } f\text{)},\\
  q &\geq 1 \quad \text{(shape exponent)}.
\end{align*}

The parametric form of the hull boundary is:

\begin{equation}
\begin{aligned}
  x(t)&=a\,\mathrm{sgn}(\cos t)\,|\cos t|^{2/q}, \\
  y(t)&=b\,\mathrm{sgn}(\sin t)\,|\sin t|^{2/q},
\end{aligned}
\quad t \in [0, 2\pi).
\label{eq:parametric}
\end{equation}

\noindent Here, $t$ traces the superellipse boundary in the hull’s local frame.  In contrast, $\theta$ denotes the hull’s orientation angle with respect to the image axes.

%
Special cases:
\begin{itemize}
  \item \(q = 1\): rhombus (diamond hull).
  \item \(q = 2\): ellipse.
  \item \(q \to \infty\): rectangle (full OBB).
\end{itemize}

\subsection{Projection onto Axis-Aligned Axes}
\begin{figure}[hbt!]
    \centering
    \includegraphics[width=\columnwidth]{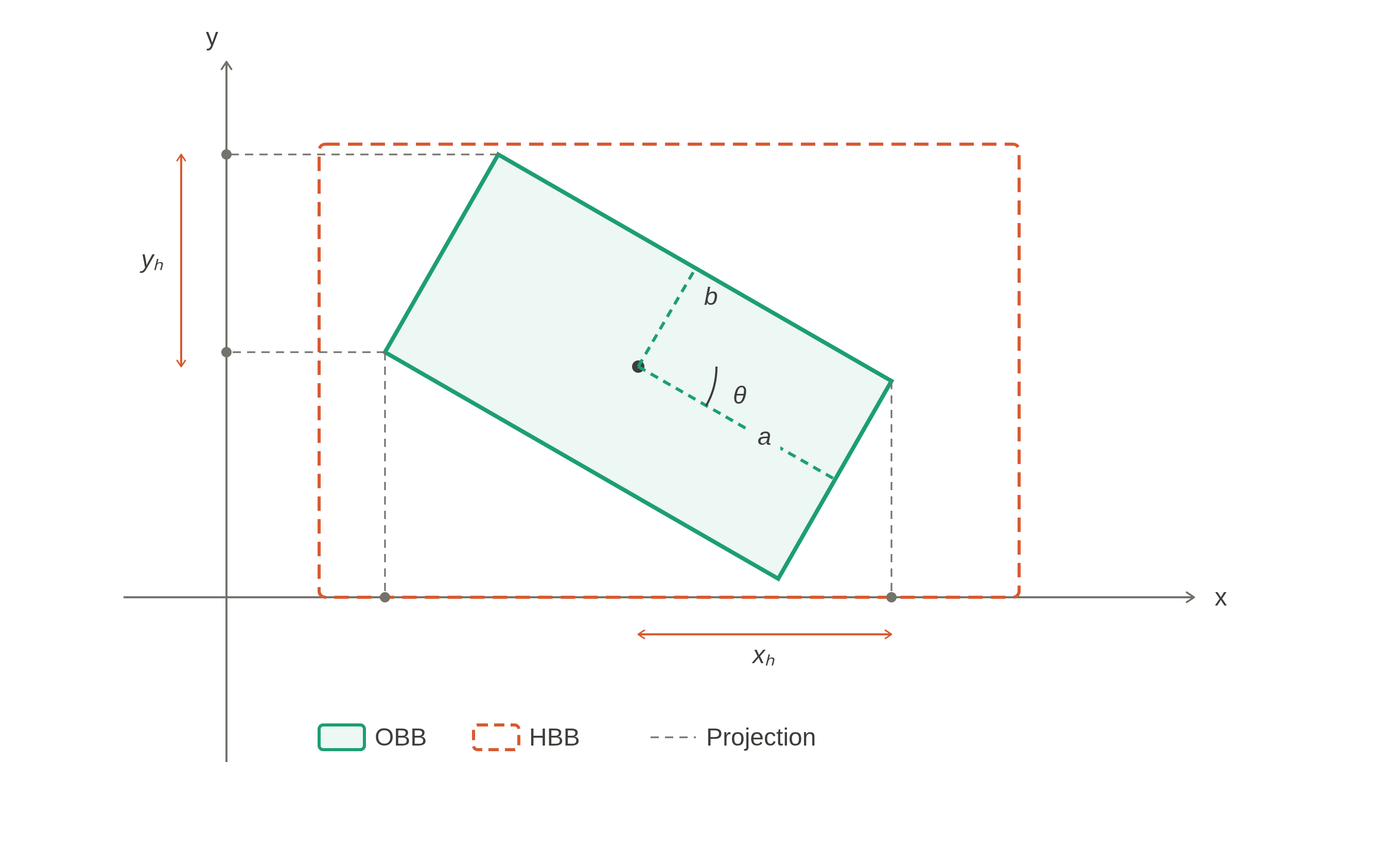}
    \caption{Illustration of OBB-HBB projection via rotation using Equations, \ref{eq:xh_max} and \ref{eq:yh_max}}
    \label{fig:obb_hbb_rotation}
\end{figure}

After rotating the hull by \(\theta\) and projecting onto the world \(x\)- and \(y\)-axes, the HBB half-extents are:
\begin{equation}
  x_h = \max_{u \in [0,1]}\!\Big[
      a\,u\,|\sin\theta| + b\,(1 - u^{q})\,|\cos\theta|
  \Big],
  \label{eq:xh_max}
\end{equation}
\begin{equation}
  y_h = \max_{v \in [0,1]}\!\Big[
      a\,v\,|\cos\theta| + b\,(1 - v^{q})\,|\sin\theta|
  \Big].
  \label{eq:yh_max}
\end{equation}

Here \(u\) and \(v\) are mixing scalars that interpolate between the \emph{corner contribution} (large \(u\)/\(v\), OBB-corner-like) and the \emph{midpoint contribution} (small \(u\)/\(v\), hull-centre-like).

\subsection{Optimal Mixing Parameters}\label{OP-MIX-PARAMs}

Let
\[
g_x(u)=a\,u\,|\sin\theta|
+b\,(1-u^q)\,|\cos\theta|.
\]

Taking the derivative,

\[
\frac{dg_x}{du}
=
a|\sin\theta|
-
bq\,u^{q-1}|\cos\theta|.
\]

Setting \(\frac{dg_x}{du}=0\) yields

\[
t_x
=
\left(
\frac{a|\sin\theta|}
     {bq|\cos\theta|}
\right)^{\frac{1}{q-1}},
\qquad \text{clipped to } [0,1].
\]

The second derivative is

\[
\frac{d^2g_x}{du^2}
=
-bq(q-1)u^{q-2}|\cos\theta|
\le 0,
\]

so \(g_x(u)\) is concave for \(q>1\). Therefore \(t_x\) is the global maximizer of \(g_x(u)\). An identical derivation for

\[
g_y(v)
=
a\,v\,|\cos\theta|
+
b\,(1-v^q)\,|\sin\theta|
\]

yields \(t_y\). Substituting \(t_x\) and \(t_y\) into
Eqs.~\eqref{eq:xh_max}--\eqref{eq:yh_max}
produces the final half-extents given in
Eqs.~\eqref{eq:xh_final}--\eqref{eq:yh_final}.

\begin{equation}
  t_x = \left(
    \frac{|\sin\theta|}{|\cos\theta|} \cdot \frac{1}{\,q \cdot b/a\,}
  \right)^{\!\frac{1}{q-1}},
  \qquad \text{clipped to } [0, 1],
  \label{eq:tx}
\end{equation}
\begin{equation}
  t_y = \left(
    \frac{|\cos\theta|}{|\sin\theta|} \cdot \frac{1}{\,q \cdot b/a\,}
  \right)^{\!\frac{1}{q-1}},
  \qquad \text{clipped to } [0, 1].
  \label{eq:ty}
\end{equation}


\subsection{Final HBB Half-Extents}
Substituting \(t_x\), \(t_y\) back into \eqref{eq:xh_max}--\eqref{eq:yh_max} and applying shrink factor \(s \in (0, 1]\):
\begin{equation}
  \boxed{x_h = s\,\Bigl[
      a\,t_x\,|\sin\theta| + b\,(1 - t_x^{\,q})\,|\cos\theta|
  \Bigr]},
  \label{eq:xh_final}
\end{equation}
\begin{equation}
  \boxed{y_h = s\,\Bigl[
      a\,t_y\,|\cos\theta| + b\,(1 - t_y^{\,q})\,|\sin\theta|
  \Bigr]}.
  \label{eq:yh_final}
\end{equation}

The output HBB is centered at the MRR centroid \((\hat{x}, \hat{y})\):
\begin{equation}
  \text{HBB} = [\hat{x} - x_h,\; \hat{x} + x_h]
             \times [\hat{y} - y_h,\; \hat{y} + y_h].
  \label{eq:hbb}
\end{equation}

The shrink factor \(s\) is a configurable parameter that scales the computed HBB half-extents. It provides a mechanism for reducing excess background area introduced during OBB-to-HBB conversion while preserving the overall geometry of the converted bounding box.

\paragraph{Degenerate cases.}
\begin{align*}
  \theta \approx 0^{\circ} &\implies t_x = 0,\; t_y = 1
      \quad \Rightarrow\; x_h = b\,s,\; y_h = a\,s,\\
  \theta \approx 90^{\circ} &\implies t_x = 1,\; t_y = 0
      \quad \Rightarrow\; x_h = a\,s,\; y_h = b\,s.
\end{align*}

\subsection{Relative Error Definition}\label{REL-ERR-DEF}
For the synthetic calibration study, we measure geometric mismatch using relative error. Given a predicted converted box with area $A_{\mathrm{pred}}$ and a reference box or target hull area $A_{\mathrm{ref}}$, the relative error is defined as
\begin{equation}
\mathrm{RelErr} = \frac{\left|A_{\mathrm{pred}} - A_{\mathrm{ref}}\right|}{A_{\mathrm{ref}}}.
\label{eq:relerr}
\end{equation}
Lower values indicate a closer geometric match between the converted box and the reference shape.

\noindent\textbf{Note:} \textit{This metric captures only area disagreement. It does not measure overlap quality, aspect-ratio mismatch, or localization fidelity in the same way as IoU. For that reason, relative error should be interpreted as a geometric calibration metric rather than a replacement for detection metrics.
}

\subsection{Match\textsubscript{50} Definition}

In addition to mean IoU, we report Match\textsubscript{50}, defined as the proportion of converted bounding boxes achieving an IoU of at least 0.5 with the reference bounding box:

\begin{equation}
\mathrm{Match}_{50}
=
\frac{\#\{\,\mathrm{IoU} \ge 0.5\,\}}
{N},
\end{equation}

where \(N\) is the total number of evaluated samples.

While mean IoU measures average geometric agreement across all samples, Match\textsubscript{50} measures the proportion of conversions that satisfy a commonly used localization threshold. This metric is useful for distinguishing methods that achieve similar mean IoU values but differ in the consistency of producing acceptable bounding boxes.

\section{Algorithm}
\label{sec:algorithm}

This section summarizes the computational procedure used to convert an oriented bounding box (OBB) into a shape-aware horizontal bounding box (HBB). The detailed derivation is given in the preceding sections; the present algorithm describes the implementation steps.

\subsection*{Input}
\begin{itemize}
  \item OBB parameters from the minimum rotated rectangle (MRR): centroid $(\hat{x}, \hat{y})$, length $L$, width $W$, and rotation angle $\theta$.
  \item Shape parameters: shape exponent $q \geq 1$, fullness factor $f \in (0,1]$, and shrink factor $s \in (0,1]$.
  \item Optional physical caps: maximum ship length $L_{\max}$ and maximum ship width $W_{\max}$.
\end{itemize}

\subsection*{Output}
\begin{itemize}
  \item Axis-aligned HBB defined by half-extents $(x_h, y_h)$ and centroid $(\hat{x}, \hat{y})$.
  \item Optional diagnostic metadata, including $L$, $W$, $\theta$, $t_x$, $t_y$, and the cap flag.
\end{itemize}

\begin{algorithm}[t]
\caption{Shape-Aware OBB-to-HBB Conversion}
\label{alg:obb2hbb}
\footnotesize
\begin{algorithmic}[1]
\Require Centroid $(\hat{x}, \hat{y})$, length $L$, width $W$, angle $\theta$
\Require Shape parameters $q$, $f$, $s$; optional caps $L_{\max}$, $W_{\max}$
\Ensure HBB $(x_{\min}, y_{\min}, x_{\max}, y_{\max})$
\State \(\varepsilon = 10^{-12}\)

\State $a \gets L/2$
\State $b \gets (W/2)\cdot f$
\State $t \gets |\theta|$
\State $s_\theta \gets \sin(t)$, $c_\theta \gets \cos(t)$
\State $s_a \gets |s_\theta|$, $c_a \gets |c_\theta|$

\If{$t < \varepsilon$}
    \State $t_x \gets 0$
    \State $t_y \gets 1$
\ElsIf{$|\pi/2 - t| < \varepsilon$}
    \State $t_x \gets 1$
    \State $t_y \gets 0$
\Else
    \State $k \gets \max\!\left(q\cdot \frac{b}{a}, \varepsilon\right)$
    \State $r_x \gets \dfrac{s_a}{\max(c_a,\varepsilon)\cdot k}$
    \State $r_y \gets \dfrac{c_a}{\max(s_a,\varepsilon)\cdot k}$
    \State $t_x \gets \min\!\left(1, \max\!\left(0, r_x^{1/(q-1)}\right)\right)$
    \State $t_y \gets \min\!\left(1, \max\!\left(0, r_y^{1/(q-1)}\right)\right)$
\EndIf

\State $x_h \gets s\cdot \bigl(a\,t_x\,s_a + b\,(1-t_x^q)\,c_a\bigr)$
\State $y_h \gets s\cdot \bigl(a\,t_y\,c_a + b\,(1-t_y^q)\,s_a\bigr)$

\If{$L_{\max}$ or $W_{\max}$ specified}
    \State $c_L \gets L_{\max}/2$
    \State $c_W \gets W_{\max}/2$
    \If{$x_h \geq y_h$}
        \State $x_h \gets \min(x_h, c_L)$
        \State $y_h \gets \min(y_h, c_W)$
    \Else
        \State $x_h \gets \min(x_h, c_W)$
        \State $y_h \gets \min(y_h, c_L)$
    \EndIf
\EndIf

\State $x_{\min} \gets \hat{x} - x_h$, \quad $x_{\max} \gets \hat{x} + x_h$
\State $y_{\min} \gets \hat{y} - y_h$, \quad $y_{\max} \gets \hat{y} + y_h$

\State \Return $(x_{\min}, y_{\min}, x_{\max}, y_{\max})$
\end{algorithmic}
\end{algorithm}

\subsection*{Numerical Stability Notes}
\begin{itemize}
  \item Degenerate orientations near $0^\circ$ and $90^\circ$ are handled explicitly to avoid division by zero.
  \item A small $\varepsilon > 0$ is used in trigonometric ratios for numerical stability.
  \item The exponent $q$ is constrained to satisfy $q \geq 1 + \varepsilon$.
  \item The parameters $t_x$ and $t_y$ are clipped to $[0,1]$ to preserve valid interpolation.
\end{itemize}

\section{Experimental Results}
\subsection{Synthetic Shape Examples}\label{SYN-SHP-EXM}
\begin{figure}[htbp]
    \centering
    \includegraphics[width=0.95\linewidth]{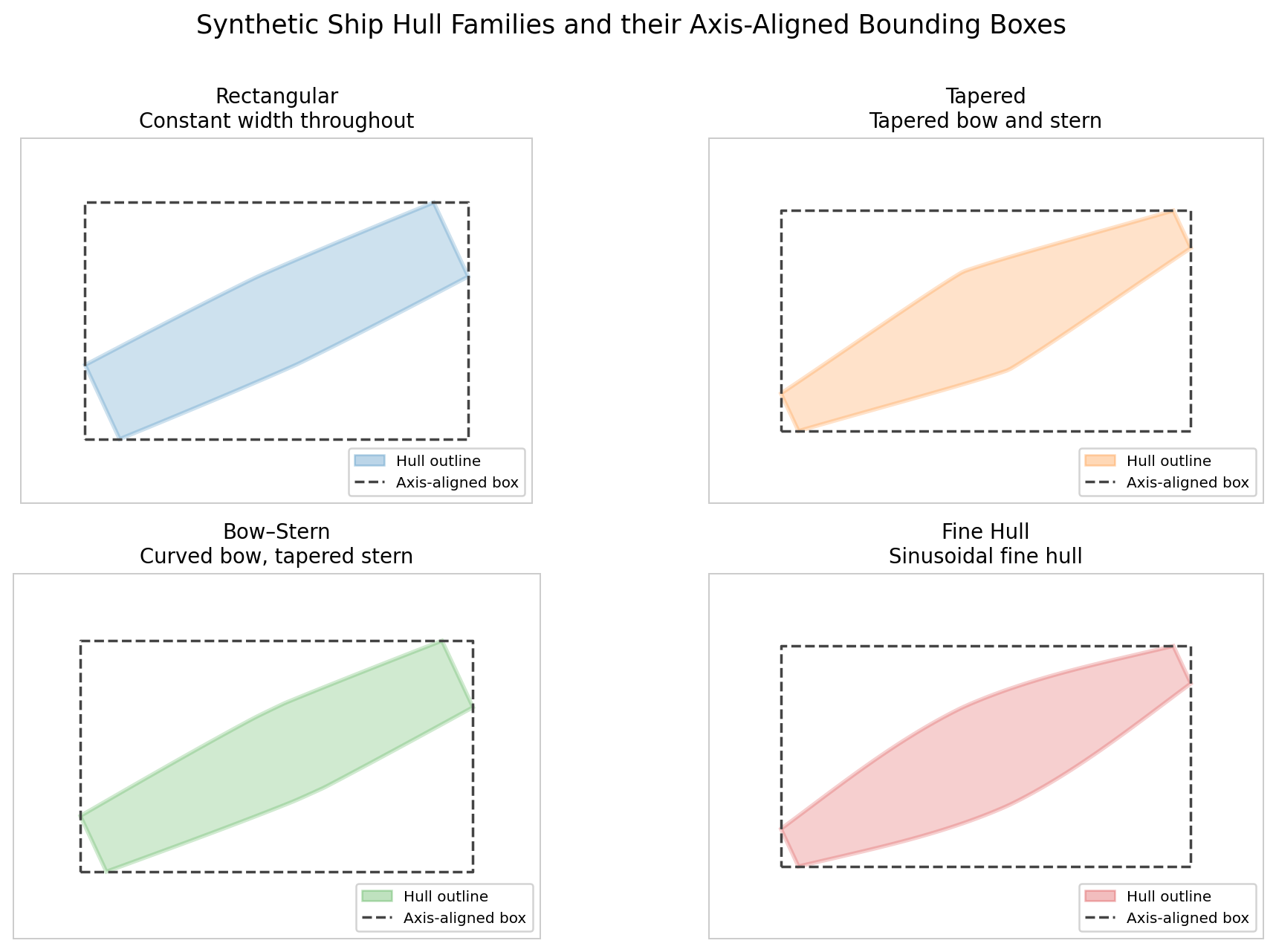}
    \caption{Representative synthetic ship families used for calibration. Each panel shows one hull family with its axis-aligned reference box.}
    \label{fig:shape_examples}
\end{figure}

We evaluate the method on four synthetic ship families (Rectangular, Tapered, Bow\_stern, Fine) across fullness levels 0.60, 0.75, 0.85, and 1.00. Figure~\ref{fig:shape_examples} shows representative hulls. The panels were designed to keep the geometry comparable while still showing clear differences in hull form.

Table~\ref{tab:family_summary_transposed} summarizes the mean relative error by hull family and fullness for OuterHBB (N), AreaEquivalentHBB (M), and NovelShapeAware (S). AreaEquivalentHBB generally performs best for tapered and fine hulls, while OuterHBB remains competitive for rectangular hulls. NovelShapeAware provides a compromise between the two and serves as the calibrated conversion model used in the real-data experiments.


For consistency with the real-data experiments, three OBB-to-HBB conversion strategies were evaluated:

\begin{itemize}
    \item \textbf{OuterHBB (N):} the conventional enclosing horizontal bounding box obtained from the minimum and maximum coordinates of the rotated rectangle.

    \item \textbf{AreaEquivalentHBB (M):} an angle-aware conversion that preserves the original OBB area while projecting the object into an axis-aligned bounding box.

    \item \textbf{NovelShapeAware (S):} the proposed superellipse-based conversion method described in Section~\ref{sec:problem}.
\end{itemize}

\begin{table}[t]
\centering
\caption{
Mean relative error (Equation~\ref{eq:relerr}) by synthetic fullness and family.
N = Naive, M = Midpoint, S = Shape-Aware.
Lower values indicate better geometric matching.
}
\label{tab:family_summary_transposed}
\renewcommand{\arraystretch}{1.15}
\setlength{\tabcolsep}{5pt}
\begin{tabular}{@{}lcccc@{}}
\toprule
Fullness & Rect & Taper & Bow\_stern & Fine \\
\midrule
0.60 -- N & 0.211 & 0.421 & 0.277 & 0.413 \\
0.60 -- M & 0.218 & 0.104 & 0.191 & 0.108 \\
0.60 -- S & 0.264 & 0.174 & 0.234 & 0.175 \\
\midrule
0.75 -- N & 0.127 & 0.376 & 0.199 & 0.366 \\
0.75 -- M & 0.267 & 0.128 & 0.234 & 0.133 \\
0.75 -- S & 0.293 & 0.187 & 0.258 & 0.188 \\
\midrule
0.85 -- N & 0.073 & 0.347 & 0.148 & 0.335 \\
0.85 -- M & 0.298 & 0.143 & 0.261 & 0.150 \\
0.85 -- S & 0.310 & 0.195 & 0.272 & 0.195 \\
\midrule
1.00 -- N & 0.000 & 0.305 & 0.077 & 0.290 \\
1.00 -- M & 0.342 & 0.167 & 0.301 & 0.175 \\
1.00 -- S & 0.333 & 0.208 & 0.290 & 0.207 \\
\bottomrule
\end{tabular}

\end{table}


The synthetic calibration sweep identified a best-performing configuration of \(q=2.2\), fullness \(=0.9\), and shrink \(=1.0\), achieving a mean IoU of 0.7931 and Match\textsubscript{50} of 0.9831 on the synthetic benchmark. These parameters were subsequently used for all real-data experiments reported in this work.

Two edge cases are important. First, the rectangular family gained less from the shape-aware rule than the tapered families, which means a single global setting is not ideal for all shapes. Second, some parameter combinations produced poor matches, which means the method needs a tuning stage before deployment.

\begin{figure}[htbp]
    \centering
    \includegraphics[width=0.9\linewidth]{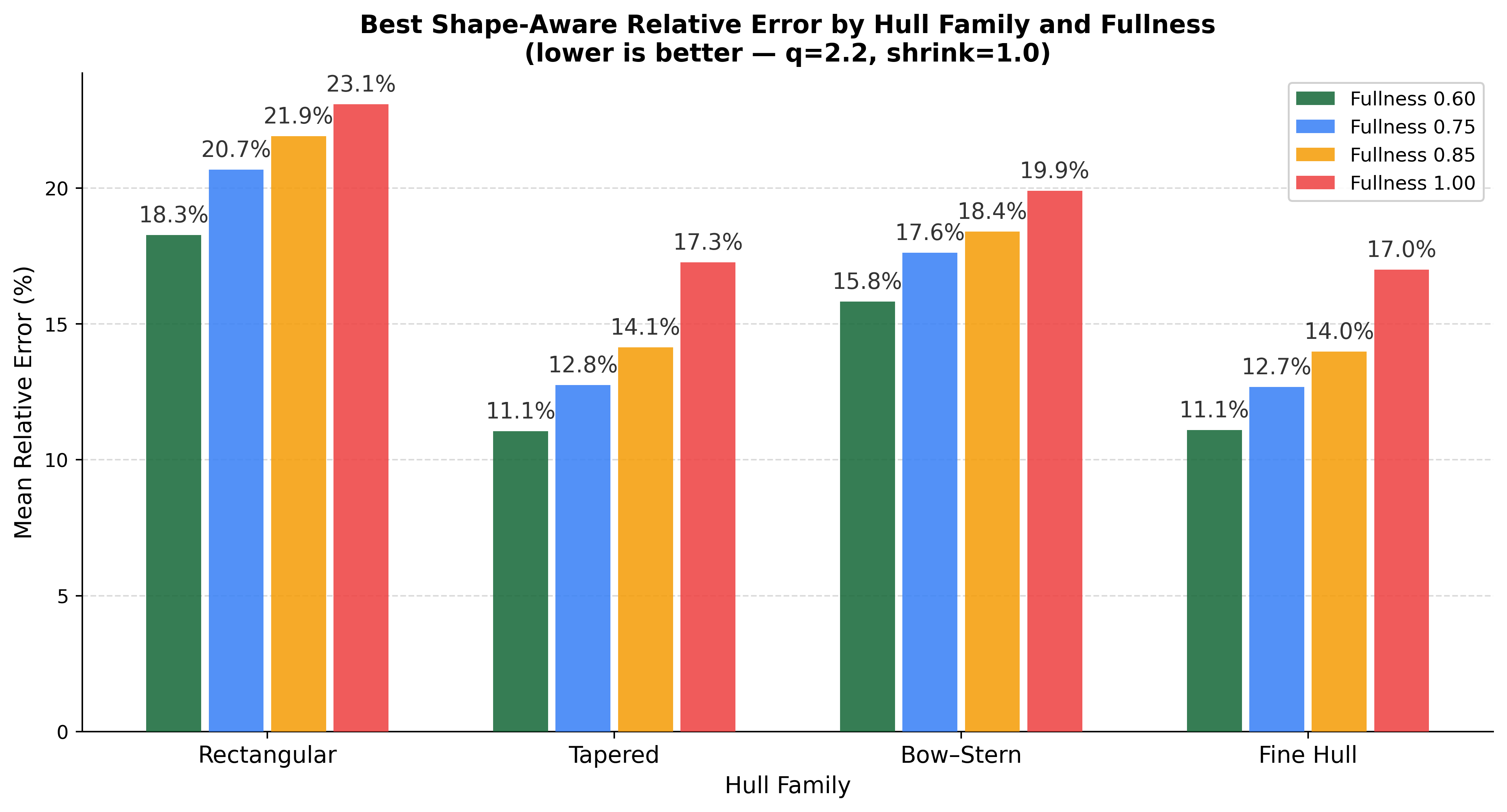}
    \caption{Best relative error achieved by the shape-aware rule for each synthetic family and fullness level.}
    \label{fig:best_error_family}
\end{figure}

\subsection{Parameter Optimization}

The shape-aware conversion introduces three tunable parameters: the superellipse exponent \(q\), the fullness factor \(f\), and the shrink factor \(s\). These parameters were optimized using a grid search on the synthetic hull dataset described in Section \ref{SYN-SHP-EXM}.

\begin{algorithm}[t]
\caption{Grid Search for Shape-Aware Parameters}
\label{alg:gridsearch}
\footnotesize
\begin{algorithmic}[1]
\Require Synthetic hull dataset \(D\)
\Require Candidate values \(Q,F,S\)
\Ensure Optimal parameters \((q^*,f^*,s^*)\)

\State bestIoU $\gets -\infty$

\ForAll{$q \in Q$}
    \ForAll{$f \in F$}
        \ForAll{$s \in S$}
            \State Convert all hulls using Algorithm~\ref{alg:obb2hbb}
            \State Compute mean IoU on \(D\)
            \If{mean IoU $>$ bestIoU}
                \State bestIoU $\gets$ mean IoU
                \State $(q^*,f^*,s^*) \gets (q,f,s)$
            \EndIf
        \EndFor
    \EndFor
\EndFor

\State \Return $(q^*,f^*,s^*)$
\end{algorithmic}
\end{algorithm}

The search space was defined as

\[
q \in \{1.5, 1.8, 2.0, 2.2, 2.4, 2.6\},
\]

\[
f \in \{0.7, 0.8, 0.9, 1.0, 1.1\},
\]

\[
s \in \{0.95, 1.0, 1.05, 1.1\}.
\]

For each combination, the converted HBB was evaluated against the
reference bounding box using mean IoU and Match\textsubscript{50}.
The parameter set maximizing mean IoU was selected as the final
configuration.

The optimal solution was found at
\(q=2.2\), \(f=0.9\), and \(s=1.0\),
yielding a mean IoU of 0.7931 and
Match\textsubscript{50}=0.9831.

\subsection{Best-Found Parameters}\label{BEST-FND-PARAM}


The optimization procedure described in Algorithm~\ref{alg:gridsearch} identified \(q=2.2\), \(f=0.9\), and \(s=1.0\) as the best-performing configuration. This parameter set achieved a mean IoU of 0.7931 and Match\textsubscript{50} of 0.9831 on the synthetic benchmark and was used for all subsequent real-data experiments.

\begin{table}[h!]
\centering
\caption{Best mixing parameters}
\label{tab:tuned_params}
\begin{tabular}{@{}lcc@{}}
\toprule
Parameter & Best (mean IoU) & Best (match\textsubscript{50}) \\
\midrule
\(q\) & 2.2 & 2.2 \\
\(f\) (fullness) & 0.9 & 0.9 \\
\(s\) (shrink) & 1.0 & 1.0 \\
\bottomrule
\end{tabular}
\end{table}


To compare the performance of each conversion method, three evaluation metrics were used: mean Intersection over Union (IoU), overshoot, and undershoot. \\
\noindent\textbf{Note:} \textit{\textbf{Overshoot} is defined as the proportion of the predicted HBB area that falls outside the original OBB, capturing how much background noise the converted box introduces. \textbf{Undershoot} is defined as the proportion of the OBB area not covered by the predicted HBB, capturing how much of the object extent the converted box fails to enclose.}

\subsection{Detection-Level Performance on ShipRSImageNet}\label{SHIP-RS-IMAGE-TUN}
The synthetic calibration results were supported by the folder-level evaluation on real world data. We benchmarked four methods on ShipRSImageNet: OuterHBB using minimum and maximum bounds of the OBB, AreaEquivalentHBB using the angle of the OBB to maintain the same area when converting to HBB, GBBMarginalized using a 2D Gaussian distribution and then marginalized across one axis for HBB, and NovelShapeAware, the paper's proposed method.\\

ShipRSImageNet is a large-scale fine-grained dataset for ship detection in high-resolution optical remote sensing images, containing over 3,435 images with 17,573 ship instances in 50 categories, annotated with both horizontal and oriented bounding boxes\cite{13.ShipRSImageNet}. Table \ref{tab:benchmark_overall} shows that NovelShapeAware achieved the highest mean IoU (0.5609) while simultaneously reducing overshoot (0.3749) and undershoot (0.1208) compared with the competing conversion methods. These results indicate a tighter geometric correspondence between the converted HBB and the underlying ship extent.


\begin{table}[h!]
\centering
\caption{Method comparison on ShipRSImageNet (10{,}043 samples).}
\label{tab:benchmark_overall}
\setlength{\tabcolsep}{2.5pt}
\renewcommand{\arraystretch}{1.15}
\begin{tabular}{lcccc}
\toprule
Method & Mean IoU & Overshoot & Undershoot & Avg. Error \\
\midrule
OuterHBB        & 0.4929 & 0.5071 & \textbf{0.0000} & 0.2536 \\
AreaEquivalentHBB  & 0.3139 & 0.5795 & 0.5736 & 0.5766 \\
GBBMarginalized & 0.2451 & 0.6909 & 0.4668 & 0.5789 \\
\textbf{NovelShapeAware} &
\textbf{0.5609} &
\textbf{0.3749} &
0.1208 &
\textbf{0.2479} \\
\bottomrule
\end{tabular}
\end{table}


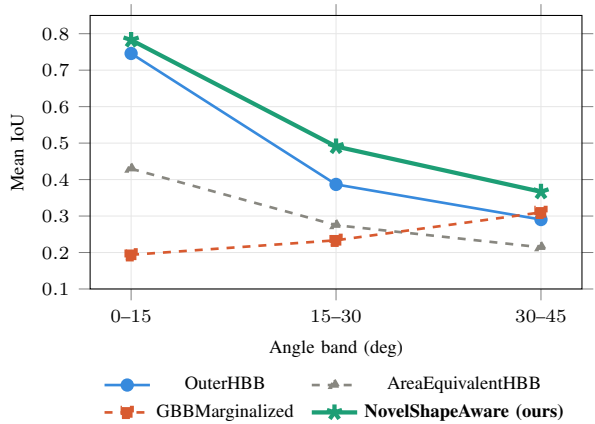
\begin{figure}[h!]
\centering
\begin{tikzpicture}
\begin{axis}[
    width=\columnwidth,          
    height=5.2cm,
    xlabel={Angle band (deg)},
    ylabel={Mean IoU},
    xlabel style={font=\scriptsize},
    ylabel style={font=\scriptsize},
    xtick={1,2,3},
    xticklabels={$0$--$15$, $15$--$30$, $30$--$45$},
    xticklabel style={font=\scriptsize},
    yticklabel style={font=\scriptsize},
    ymin=0.10, ymax=0.85,
    ytick={0.1,0.2,0.3,0.4,0.5,0.6,0.7,0.8},
    grid=both,
    grid style={line width=0.3pt, draw=gray!20},
    legend style={
        at={(0.5,-0.28)},        
        anchor=north,
        legend columns=2,
        font=\scriptsize,
        draw=none,
        fill=none,
        /tikz/every even column/.append style={column sep=0.4em},
    },
    tick align=outside,
    every axis plot/.append style={line width=1pt, mark size=2pt},
]

\addplot[color=outerHBB,  mark=*,         solid]
    coordinates {(1,0.7458)(2,0.3867)(3,0.2904)};
\addlegendentry{OuterHBB}

\addplot[color=areaEquiv, mark=triangle*,  dashed]
    coordinates {(1,0.4297)(2,0.2749)(3,0.2143)};
\addlegendentry{AreaEquivalentHBB}

\addplot[color=gbbMarg,   mark=square*,   dashed]
    coordinates {(1,0.1937)(2,0.2335)(3,0.3100)};
\addlegendentry{GBBMarginalized}

\addplot[color=novelShape, mark=star, solid, line width=1.6pt, mark size=3pt]
    coordinates {(1,0.7828)(2,0.4908)(3,0.3666)};
\addlegendentry{\textbf{NovelShapeAware (ours)}}

\end{axis}
\end{tikzpicture}
\caption{Angle-stratified mean IoU on ShipRSImageNet (10,043 samples).
\textbf{NovelShapeAware} demonstrates superior robustness to object rotation.}
\label{fig:angle_iou_linegraph}
\end{figure}



For a better understanding of how NovelShapeAware performs at different orientation angles, we calculated the mean IoU across three different angle bands, \(0^\circ\)- \(15^\circ\), \(15^\circ\)- \(30^\circ\), \(30^\circ\)- \(45^\circ\). These three bands span the full range of unique orientations, as any orientation angle beyond \(45^\circ\) is geometrically equivalent to a reflection about one of the image axis. Figure \ref{fig:angle_iou_linegraph} shows the mean IoU split across the three different angle bands. Across all three angle bands, NovelShapeAware achieved the highest mean IoU, obtaining 0.7828 for 0$^\circ$--15$^\circ$, 0.4908 for 15$^\circ$--30$^\circ$, and 0.3666 for 30$^\circ$--45$^\circ$. This indicates improved robustness to object rotation compared with the alternative conversion approaches.

\subsection{Performance on Sentinel 2 Data}\label{S2-CUST-DATA}

We collected data from Sentinel 2 containing ground truth for OBB and HBB. This dataset was used as a real-world test for the proposed data on unseen data. The collected data contained 66 samples with 35 candidates in \(0^\circ\)- \(15^\circ\), 11 candidates in \(15^\circ\)- \(30^\circ\) and 20 candidates in \(30^\circ\)- \(45^\circ\). Table \ref{tab:s2_benchmark} provides the mean IoU, overshoot and undershoot between the proposed method and the three standards, OuterHBB, AreaEquivalentHBB, and GBBMarginalized. With this custom dataset, our method, NovelShapeAware performs better with higher mean IoU score and lower overshoot scores. The lower undershoot for OuterHBB compared to NovelShapeAware occured due to oversized HBB which sometimes also included the wake of the ship.

\begin{table}[h!]
\centering
\caption{Method comparison on Sentinel 2 dataset.}
\label{tab:s2_benchmark}
\setlength{\tabcolsep}{2.5pt}
\renewcommand{\arraystretch}{1.15}
\begin{tabular}{lcccc}
\toprule
Method & Mean IoU & Overshoot & Undershoot \\
\midrule
OuterHBB        & 0.2871 & 0.673 & \textbf{0.2139} \\
AreaEquivalentHBB  & 0.2662 & 0.6435 & 0.3895 \\
GBBMarginalized & 0.2299 & 0.697 & 0.4204 \\
\textbf{NovelShapeAware} &
\textbf{0.3527} &
\textbf{0.5217} &
0.3562 \\
\bottomrule
\end{tabular}
\end{table}

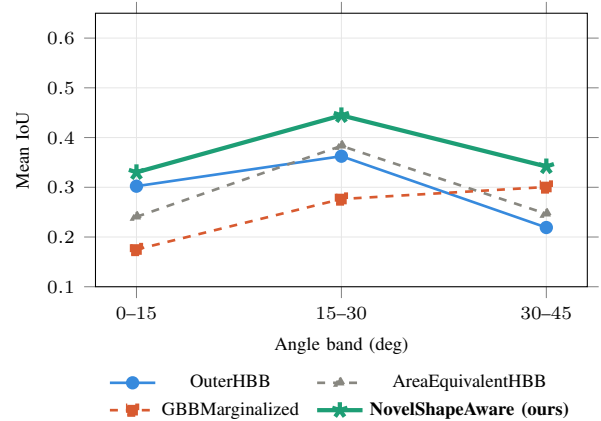
\begin{figure}[hbt!]
\centering
\begin{tikzpicture}
\begin{axis}[
    width=\columnwidth,          
    height=5.2cm,
    xlabel={Angle band (deg)},
    ylabel={Mean IoU},
    xlabel style={font=\scriptsize},
    ylabel style={font=\scriptsize},
    xtick={1,2,3},
    xticklabels={$0$--$15$, $15$--$30$, $30$--$45$},
    xticklabel style={font=\scriptsize},
    yticklabel style={font=\scriptsize},
    ymin=0.10, ymax=0.65,
    ytick={0.1,0.2,0.3,0.4,0.5,0.6},
    grid=both,
    grid style={line width=0.3pt, draw=gray!20},
    legend style={
        at={(0.5,-0.28)},        
        anchor=north,
        legend columns=2,
        font=\scriptsize,
        draw=none,
        fill=none,
        /tikz/every even column/.append style={column sep=0.4em},
    },
    tick align=outside,
    every axis plot/.append style={line width=1pt, mark size=2pt},
]

\addplot[color=outerHBB,  mark=*,         solid]
    coordinates {(1,0.3022)(2,0.3625)(3,0.2191)};
\addlegendentry{OuterHBB}

\addplot[color=areaEquiv, mark=triangle*,  dashed]
    coordinates {(1,0.2406)(2,0.3832)(3,0.2466)};
\addlegendentry{AreaEquivalentHBB}

\addplot[color=gbbMarg,   mark=square*,   dashed]
    coordinates {(1,0.1749)(2,0.2764)(3,0.3006)};
\addlegendentry{GBBMarginalized}

\addplot[color=novelShape, mark=star, solid, line width=1.6pt, mark size=3pt]
    coordinates {(1,0.3301)(2,0.4446)(3,0.3418)};
\addlegendentry{\textbf{NovelShapeAware (ours)}}

\end{axis}
\end{tikzpicture}
\caption{Angle-stratified mean IoU on Sentinel 2 Dataset.}
\label{fig:s2-angle_iou_linegraph}
\end{figure}

Figure \ref{fig:s2-angle_iou_linegraph} provides a visualization for the IoU split between the three angle bands. Across all three the proposed method performed better than the ccommonly used conversion methods. 

\begin{figure}[h!]
    \centering
    \begin{subfigure}{\columnwidth}
        \centering
        \begin{minipage}[t]{0.45\columnwidth}
            \centering
            \includegraphics[width=\linewidth]{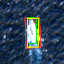}
        \end{minipage}
        \hfill
        \begin{minipage}[t]{0.45\columnwidth}
            \centering
            \includegraphics[width=\linewidth]{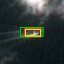}
        \end{minipage}
        \caption{Angle band \(0^\circ\)- \(10^\circ\): low-rotation detections.}
        \label{fig:row1}
    \end{subfigure}

    \vspace{0.3em}

    \begin{subfigure}{\columnwidth}
        \centering
        \begin{minipage}[t]{0.45\columnwidth}
            \centering
            \includegraphics[width=\linewidth]{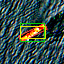}
        \end{minipage}
        \hfill
        \begin{minipage}[t]{0.45\columnwidth}
            \centering
            \includegraphics[width=\linewidth]{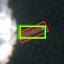}
        \end{minipage}
        \caption{Angle band \(15^\circ\)- \(30^\circ\): moderate-rotation detections.}
        \label{fig:row2}
    \end{subfigure}

    \vspace{0.3em}

    \begin{subfigure}{\columnwidth}
        \centering
        \begin{minipage}[t]{0.45\columnwidth}
            \centering
            \includegraphics[width=\linewidth]{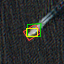}
        \end{minipage}
        \hfill
        \begin{minipage}[t]{0.45\columnwidth}
            \centering
            \includegraphics[width=\linewidth]{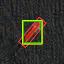}
        \end{minipage}
        \caption{Angle band \(30^\circ\)- \(45^\circ\): high-rotation detections.}
        \label{fig:row3}
    \end{subfigure}

    \caption{Showcasing NovelShapeAware HBB (yellow) with Ground truth for OBB (Red) and HBB (Green)}
    \label{fig:S2-showcase}
\end{figure}

Figure \ref{fig:S2-showcase} shows images with the ground truth OBB in red, ground truth HBB in green and the HBB generated using the proposed method in yellow. As it can be observed, the proposed method provides a tighter and more aligned fit for the bounding box.

\section{Observations}

The synthetic experiments demonstrated that conversion performance varies across hull families. Rectangular hulls generally exhibited higher relative errors, whereas tapered and fine hulls were associated with lower errors. This behaviour indicates that the suitability of a particular conversion strategy depends on the underlying hull geometry and supports the use of parameter calibration prior to deployment.

The synthetic calibration study identified (q=2.2), fullness (=0.9), and shrink (=1.0) as the best-performing parameter configuration. These values were subsequently adopted for the real-data evaluation.

On ShipRSImageNet, the proposed shape-aware method achieved the highest mean IoU among the evaluated conversion approaches (0.5609), while also reducing overshoot (0.3749) and undershoot (0.1208) relative to the competing methods. Performance remained robust across all evaluated orientation ranges, achieving mean IoUs of 0.7828, 0.4908, and 0.3666 for the 0$^\circ$--15$^\circ$, 15$^\circ$--30$^\circ$, and 30$^\circ$--45$^\circ$ angle bands, respectively.

\section{Edge-Case Analysis}
Two edge cases are important. First, the rectangular family gained less from the shape-aware rule than the tapered families, which means a single global setting is not ideal for all shapes. Second, some parameter combinations produced poor matches, which means the method needs a tuning stage before deployment.

The maximum observed shape-aware relative error was 0.6689, which confirms that the method is sensitive to parameter choice and should not be used without calibration.

Degenerate orientation cases are handled by explicit clipping:
\begin{itemize}
  \item \(\theta \approx 0^\circ\): \(t_x = 0, t_y = 1\), yielding \(x_h = b\,s, y_h = a\,s\).
  \item \(\theta \approx 90^\circ\): \(t_x = 1, t_y = 0\), yielding \(x_h = a\,s, y_h = b\,s\).
\end{itemize}

These cases are robustly handled by the algorithm through clipping and conditional checks for \(\sin\theta \approx 0\) or \(\cos\theta \approx 0\).

\section{Discussion}
The main takeaway is that the proposed converter should be viewed as a calibrated rule rather than a fixed default. The synthetic experiments show where the method works best, the edge cases show where it underperforms, and the real-data evaluation shows that the tuned configuration transfers into detection gains. This supports the use of synthetic calibration as a necessary step before applying the method to real annotations.

The shape-aware method reduces both overshoot and undershoot relative to OuterHBB and AreaEquivalentHBB, while maintaining higher IoU. This indicates that the super-ellipse based hull model captures real ship geometry more accurately than box-corner or Gaussian-based approximations.

\section{Conclusion}



The paper proposes a shape aware OBB to HBB conversion method based on a super-ellipse hull model with three tunable parameters. The method using relative error defined in Section \ref{REL-ERR-DEF} derives the optimal mixing parameters for axis-aligned projection and provides a stable algorithm with degenerate-case handling.

Through comparison between NovelShapeAware and the three commonly used conversion methods on two datasets, ShipRSImageNet and a custom Sentinel 2 dataset, the paper demonstrates a superior conversion method with a higher intersection over union towards different orientation angles. The method also provides tighter HBB compared to other methods as illustrated by higher mean IoU scores in both ShipRSImageNet and our custom Sentinel 2 dataset.

\section{Future Work}
Future work includes:
\begin{itemize}
  \item Validating shape exponent \(q\) across additional ship classes and datasets.
  \item Analyzing sensitivity of fullness factor \(f\) on narrow vs wide hulls.
  \item Investigating adaptive \(q, f\) based on aspect ratio \(L/W\).
  \item Adding visualization plots for OBB vs shape-aware HBB.
\end{itemize}



\bibliographystyle{IEEEtran}
\bibliography{references}

\end{document}